\documentclass[letterpaper, 10 pt, conference]{ieeeconf}  

\IEEEoverridecommandlockouts                              

\usepackage{tikz}
\usetikzlibrary{arrows.meta, positioning}
\usepackage{graphicx}
\usepackage{booktabs}

\title{\LARGE \bf
Surgical Re-enactment for Operating Room Workflow Datasets*}

\author{Jana Nina Friedrich$^{1}$, Andrea Karin Maria Ross$^{2}$, Angelo Henriques$^{1}$, \\ Mario Peter Martin Weisser$^{1}$, Ling Zhang$^{3}$ and Mohammad Ali Nasseri$^{1,4}$%
\thanks{*This work is supported by the BMFTR under the funding directive for Natural Language Integration of Robotics in Healthcare Facilities (GRATA project).}%
\thanks{$^{1}$J. N. Friedrich, A. Henriques, M. P. M. Weisser and M. A. Nasseri are with the Medical Autonomy and Precision Surgery Lab, TUM University Hospital, Munich, Germany.
        {\tt\small jana.friedrich@tum.de, angelo.henriques@tum.de, mario.weisser@tum.de, ali.nasseri@tum.de}}%
\thanks{$^{2}$A. K. M. Ross is with the TUM School of Medicine and Health, Department of Ophthalmology, TUM University Hospital, Munich, Germany.
        {\tt\small andrea.ross@mri.tum.de}}%
\thanks{$^{3}$L. Zhang is with the TUM School of Computation, Information, and Technology, TUM, Munich, Germany.
        {\tt\small ling.zhang@tum.de}}%
\thanks{$^{4}$M. A. Nasseri is also with the Department of Biomedical Engineering, University of Alberta, Edmonton, Canada.}%
}
 
\begin{document}
 
\maketitle
\thispagestyle{empty}
\pagestyle{empty}
 

\begin{abstract}
The introduction of new technologies, such as surgical robots, is driving the vision of a connected, smart operating room (OR). However, realizing this vision requires a deep understanding of surgical workflows, which relies on realistic datasets capturing the actions of all OR personnel from both full room and surgical field perspectives. Acquiring such data in real ORs is prohibitively challenging due to factors such as ethics committee approvals, limited space for camera installation, and sterility regulations preventing the use of tracking markers. We present a step-by-step methodology for re-enacting complete surgical procedures in a reconstructed OR. This approach enables the creation of repeatable and annotatable workflow datasets for training activity recognition models, generating scene graphs, and formalizing surgical process models. Developed for robot-assisted ophthalmic surgery, our methodology combines expert consultation, structured workflow formalization, OR reconstruction, role-based training, real OR observation, and iterative recording with post-take debriefing. We provide concrete recommendations to allow other research groups to seamlessly adopt this methodology for their own surgical domains.
\end{abstract}
 
\section{INTRODUCTION}
 
The growing adoption of surgical robots and networked devices is transforming operating rooms (ORs) into connected, context-aware environments. A key requirement for such systems is understanding surgical workflows. A surgical workflow is a coordinated process carried out by various OR personnel. Some tasks depend on preceding steps by other OR staff, creating complex dependencies. AI-based training and assistance systems need to understand these dependencies, which requires datasets that capture the complete workflow of all personnel, not just the surgeon's actions. Workflows can be represented by formal frameworks such as  surgical process modeling~\cite{neumuth2017}, workflow recognition~\cite{neumann2022}, and Business Process Model and Notation (BPMN)~\cite{neumann2019}. However, they depend on annotatable data from realistic surgical environments.
 
Acquiring this data in real clinical settings is prohibitively difficult. Ethical approvals, sterility requirements, space constraints, and suboptimal lighting all limit the feasibility of multi-camera recording in active ORs. Existing surgical video datasets such as Cholec80~\cite{cholec80} capture only endoscopic views, not the full OR scene. Sharghi et al.~\cite{sharghi2020} collected 400 multi-perspective OR videos during real robot-assisted surgeries, but such institutional access is rare. The 4D-OR dataset~\cite{ozsoy2022} uses simulated knee replacement surgeries for scene graph generation, but does not describe the re-enactment methodology, making it difficult for other groups to reproduce.
 
Unlike existing works, we present an eight-phase reproducible methodology for re-enacting complete surgical procedures in a reconstructed OR. We developed this approach as part of an AI-based training platform combining semantic knowledge models, large language models, and perception. While designed for robot-assisted ophthalmic surgery~\cite{nasseri2013}, the methodology is transferable to other surgical domains.
 
\section{METHODOLOGY}
 
\begin{figure}[t]
\centering
\resizebox{\columnwidth}{!}{%
\begin{tikzpicture}[
    node distance=0.35cm and 0.25cm,
    box/.style={rectangle, draw, rounded corners=2pt, text centered, font=\scriptsize, minimum height=0.8cm, text width=1.7cm, thick},
    wfbox/.style={box, fill=blue!15},
    rehbox/.style={box, fill=green!15},
    defbox/.style={box, fill=white},
    arr/.style={-{Stealth[length=2mm]}, thick},
    feedback/.style={-{Stealth[length=2mm]}, thick, dashed}
]
 
\node[defbox] (p1) {1. Expert Consultation};
\node[wfbox, right=of p1] (p2) {2. Initial Workflow};
\node[defbox, right=of p2] (p3) {3. Role Assignment};
\node[defbox, right=of p3] (p4) {4. OR Reconstruction};
 
\node[rehbox, below=0.6cm of p4] (p5) {5. Role Training};
\node[defbox, left=of p5] (p6) {6. OR Observation};
 
\node[box, left=of p6, fill=white] (p7) {7. Final Preparation};
\begin{scope}
\clip (p7.north west) rectangle (p7.south);
\fill[blue!15] (p7.north west) rectangle (p7.south east);
\end{scope}
\begin{scope}
\clip (p7.north) rectangle (p7.south east);
\fill[green!15] (p7.north west) rectangle (p7.south east);
\end{scope}
\draw[thick, rounded corners=2pt] (p7.north west) rectangle (p7.south east);
\node[font=\scriptsize, text centered, text width=1.7cm] at (p7) {7. Final Preparation};
 
\node[defbox, left=of p7] (p8) {8. Recording};
 
\draw[arr] (p1) -- (p2);
\draw[arr] (p2) -- (p3);
\draw[arr] (p3) -- (p4);
\draw[arr] (p4) -- (p5);
\draw[arr] (p5) -- (p6);
\draw[arr] (p6) -- (p7);
\draw[arr] (p7) -- (p8);
 
\draw[feedback] ([xshift=-0.5cm]p8.north) -- ++(0,0.22) -- ++(1.0,0) node[above, font=\tiny, pos=0.5] {Debrief} -- ++(0,-0.26);
 
\node[fill=blue!15, draw, rounded corners=2pt, inner sep=2pt, font=\tiny, minimum height=0.35cm, anchor=north east] at ([xshift=-1.35cm]p8.north west) (lw) {Workflow};
\node[fill=green!15, draw, rounded corners=2pt, inner sep=2pt, font=\tiny, minimum height=0.35cm, below=0.08cm of lw] (lr) {Rehearsal};
 
\end{tikzpicture}%
}
\vspace{-0.35cm}
\caption{The eight phases of the re-enactment methodology, progressing clockwise from expert consultation to recording. The clinical expert is involved in all phases.}
\label{fig:workflow}
\vspace{-0.4cm}
\end{figure}
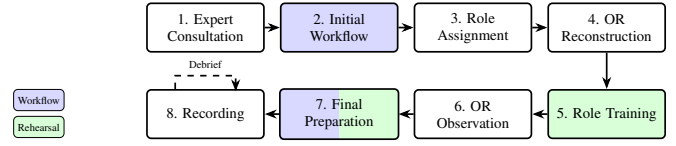

Our methodology comprises eight phases (Fig.~\ref{fig:workflow}), progressing from initial expert consultation through iterative workflow refinement to final recording.
 
\textbf{Phase 1: Expert Consultation.}
The first step is to recruit a clinical domain expert, e.g. an experienced surgeon. This expert provides an initial walkthrough of the entire OR procedure, covering preparation, surgical execution, and cleanup. It is essential that this expert guides the team in all phases, as domain-specific questions arise continuously and can only be resolved with clinical knowledge.
 
\textbf{Phase 2: Initial Workflow.}
Based on the expert walkthrough, the team creates a first draft of the surgical workflow as a structured document. This includes: (a)~a complete equipment list (instruments, machines, consumables), (b)~the OR professions the team wants to capture, and (c)~all actions in chronological order. In the context of the re-enactment, we refer to the assigned professions as roles. It is important that the workflow captures the parallelism of actions across roles, as multiple professions often act simultaneously.
 
\textbf{Phase 3: Role Assignment.}
Each team member is assigned a role to re-enact, such as surgeon, surgical assistant, anesthesiologist, runner, or cleaning staff. We additionally define a dedicated director role, not present in real ORs, but essential for dataset creation. The director does not participate in the re-enactment but observes each session and documents: (a)~deviations from the workflow, (b)~errors or hesitations, and (c)~other notable events. This documentation greatly simplifies later dataset annotation and enables the iterative debrief loop in Phase~8.
 
\textbf{Phase 4: OR Reconstruction.}
The team visits the real OR after operating hours to measure the room layout and document equipment positions, producing a floor plan. Based on this, the team sets up a reconstructed OR equipped with appropriate data-capturing systems (e.g., RGB-D cameras, microscope camera, microphones).
 
\textbf{Phase 5: Role Training.}
This phase focuses on individual practice rather than collective rehearsal. Each team member trains their assigned role separately under the guidance of the clinical expert, who provides real-time corrections.
 
\textbf{Phase 6: OR Observation.}
Each team member attends live surgeries to observe the real clinical environment. Before the visits, the team prepares a list of open questions based on uncertainties from role training. Each person shadows their assigned role and takes structured notes.
 
\textbf{Phase 7: Final Preparation.}
The observations from Phase~6 are used to update the workflow from Phase~2 to close existing gaps by adding missing steps, timing corrections, and communication patterns. The team then performs a complete end-to-end run-through of the refined workflow before proceeding to recording.
 
\begin{figure}[t]
\centering
\includegraphics[width=0.95\columnwidth]{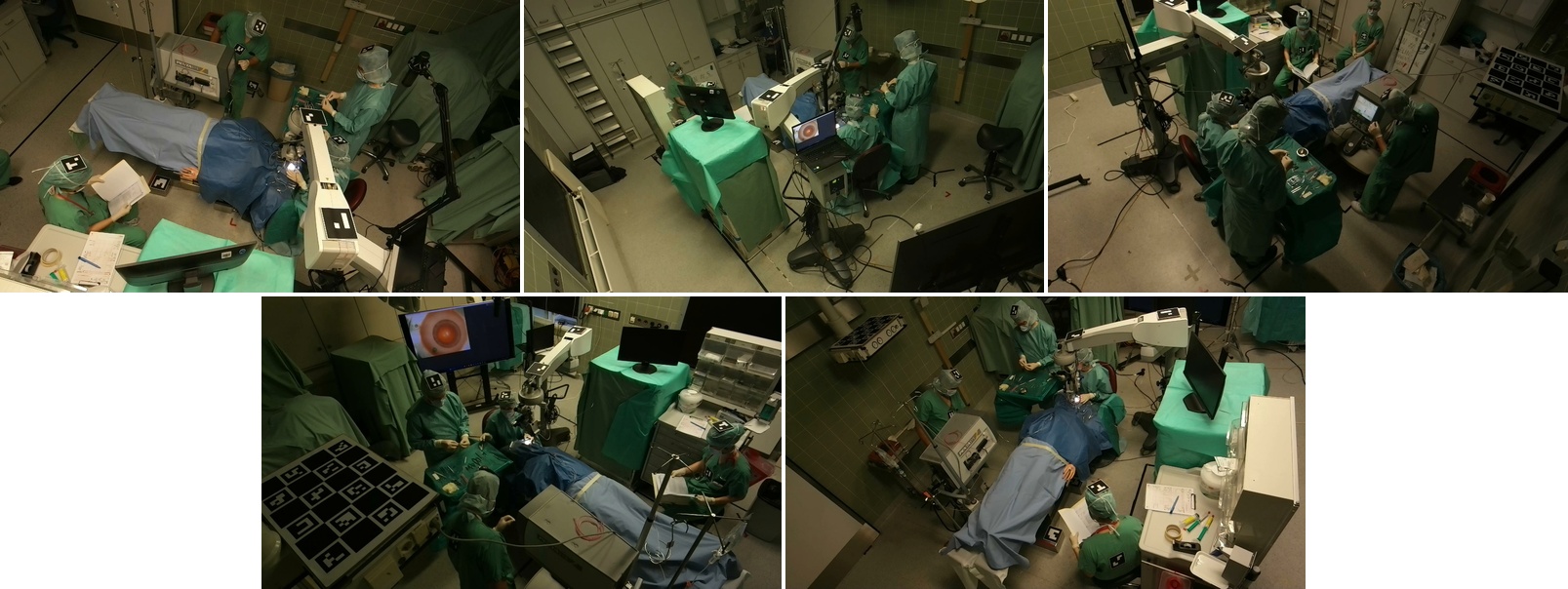}
\vspace{-0.3cm}
\caption{Synchronized frame from the five RGB-D camera perspectives during a re-enacted ophthalmic procedure.}
\label{fig:views}
\vspace{-0.4cm}
\end{figure}

\textbf{Phase 8: Recording.}
Once the clinical expert confirms that the team can perform the procedure consistently, recording begins using the data capturing systems from Phase~4 (Fig.~\ref{fig:views}). After each take, the director's notes and clinical expert's observations are discussed to identify and correct errors. This feedback loop enables rapid improvement from take to take.
 
\section{RESULTS}
 
Over multiple recording days, 10 complete re-enactments of an ophthalmic surgical procedure were captured. Each recording involved six people covering seven roles as described in Phase~3, with one person performing two non-overlapping roles. The duration ranged from 30 to 44 minutes, with early takes being longer due to timing errors, corrections of missed steps, and issues such as incorrect instrument handling or breaking the sterile field. The quality improved substantially through iterative debriefing so that later takes became more efficient and required fewer corrections. The clinical expert qualitatively assessed that the final recordings closely resembled real OR dynamics. To increase dataset diversity, selected takes included deliberately introduced variations such as a dropped instrument. The resulting dataset enables training of activity recognition models, generation of scene graphs, and formalization of BPMN-based surgical process models.
 
\section{DISCUSSION}
Compared to recording in real ORs, re-enactment offers full control over sensor placement without sterility constraints, unlimited repeatability, and the ability to introduce controlled variations. Moreover, annotation is significantly simplified through the director's log and the structured workflow document as ground truth. Only a single clinical expert is required rather than an entire OR team. Embedding this expert in the research team throughout all phases proved critical, as domain-specific questions arose continuously. Although developed for ophthalmic surgery, the methodology is transferable to other surgical domains.
 
A limitation is that the re-enactments were performed by researchers rather than real OR staff. Involving a full clinical team would increase fidelity but is difficult to coordinate due to limited availability and scheduling constraints of OR personnel. Nevertheless, for training activity recognition and workflow models, the spatial and temporal structure is more important than clinical precision of individual gestures. Beyond workflow modeling, the resulting data can also support AI-based analysis of how surgical robots can be effectively integrated into OR workflows before first real-time clinical applications are realized. Quantitative validation, e.g. using Dynamic Time Warping to compare against real OR logs, remains future work. Finally, the full annotated dataset is being prepared for public release.
 

\end{document}